\documentclass[12pt]{article}
\usepackage[utf8]{inputenc}
\usepackage[T1]{fontenc}
\usepackage[english]{babel}
\usepackage{ifpdf,newtxtext,newtxmath}
\usepackage{array,graphicx,dcolumn,multirow,hevea,abstract,hanging,fancyhdr,float}
\newcommand{\jref}{http://journal.sjdm.org/vol16.1.html}
\newcommand{\jhead}{Open publication}

\fancypagestyle{firstpage}{%
 \lhead{\protect\small \jhead}
 \rhead{}
}
\usepackage[labelfont=sc,textfont=sf]{caption}
\usepackage[hyperfootnotes=false,breaklinks=true]{hyperref} 
\usepackage[hyphenbreaks]{breakurl}

\usepackage{booktabs} 
\newcolumntype{L}[1]{>{\raggedright\arraybackslash }p{#1}} 
\newcolumntype{C}[1]{>{\centering\arraybackslash }p{#1}}
\newcolumntype{R}[1]{>{\raggedleft\arraybackslash }p{#1}}
\newcolumntype{d}[1]{D{.}{.}{#1}} 

\let\tempone\itemize
\let\temptwo\enditemize
\let\tempthree\enumerate
\let\tempfour\endenumerate
\renewenvironment{itemize}{\tempone\setlength{\itemsep}{0pt}}{\temptwo}
\renewenvironment{enumerate}{\tempthree\setlength{\itemsep}{0pt}}{\tempfour}

\title{Learning Hierarchical Integration of Foveal
and Peripheral Vision for Vergence Control by
Active Efficient Coding}

\author{Zhetuo Zhao\thanks{Dept. of Brain and Cognitive Science, University of Rochester, Rochester, NY, USA} \and
Jochen Triesch\thanks{Frankfurt Institute for Advanced Studies, Frankfurt am Main, Germany} \and
Bertram E. Shi\thanks{Dept. of Electronic and Computer Engineering,
Hong Kong University of Science and Technology, Kowloon, Hong Kong}}

\date{} 
\begin{document} 


\maketitle
\thispagestyle{firstpage}

\begin{abstract}

The active efficient coding (AEC) framework parsimoniously explains the joint development of visual processing and eye movements, e.g., the emergence of binocular disparity selective neurons and fusional vergence, the disjunctive eye movements that align left and right eye images. Vergence can be driven by information in both the fovea and periphery, which play complementary roles. The high resolution fovea can drive precise short range movements. The lower resolution periphery supports coarser long range movements. The fovea and periphery may also contain conflicting information, e.g. due to objects at different depths.
While past AEC models did integrate peripheral and foveal information, they did not explicitly take into account these characteristics.
We propose here a two-level hierarchical approach that does.
The bottom level generates different vergence actions from foveal and peripheral regions. The top level selects one. We demonstrate that the hierarchical approach performs better than prior approaches in realistic environments, exhibiting better alignment and less oscillation.

\end{abstract}

\setlength{\baselineskip}{16pt plus.2pt}

\section{Introduction}

The perception-action cycle is at play during vergence eye movements.
Depth information is encoded as the population responses of
binocular disparity selective simple and complex cells in the primary visual cortex \cite{Ohzawa1990,Hubel1959,Hubel1962}.
The population responses of these neurons drive the eye muscles via the ocular motor neurons so that the two eyes converge or diverge
to align the left and right foveal images.

The fovea and the periphery play complementary roles in
vergence control.
Foveal disparity selective cells have small receptive field sizes \cite{Freeman2011}
and high preferred spatial frequencies \cite{Henriksson2008}.
They provide precise disparity detection over small ranges
\cite{Rawlings1969}.
Peripheral disparity selective cells
have larger receptive field sizes
and lower preferred spatial frequencies.
They provide robust, but less precise detection,
over larger disparity ranges.

Fusional vergence can zero out initial retinal disparities
of up to four degrees \cite{Antona2008,Stevenson1999},
while the fovea covers a region only one degree in diameter.
This suggests that fusional vergence control requires cooperation between
the fovea and periphery.
Information in the periphery is useful during the initial stages of vergence, when disparities may be large.
On the other hand,
when the foveal images are aligned,
objects located at other depths
may result in nonzero disparities in the periphery,
suggesting that some information
in the periphery should be ignored.
Tanimoto showed that disturbances presented in the periphery
influenced the vergence latency, but not the
steady state amplitude of vergence angle \cite{Tanimoto2004},
suggesting that peripheral vision is involved
in the early stages of vergence,
but not at steady state.

This paper proposes a mechanism that enables an agent
to learn how to both process and integrate visual information from the fovea and periphery in order to achieve robust vergence control.
Much past work in neuromorphic vergence control either did not address
the problem of learning and relied upon hand crafted
perceptual and control strategies
\cite{Siebert1992,Westelius1995,Gibaldi2010,Zhang2011},
or studied learning of only one aspect,
i.e. control \cite{Piater1999,Gibaldi2012,Gibaldi2015}.

The mechanism is based upon the active efficient coding (AEC) framework,
which extends the efficient coding hypothesis \cite{Barlow1961,Blattler2011} to include action.
In contrast to past work on joint learning of vergence \cite{Franz2007},
the AEC does not require a fixed pre-defined reward signal.
Prior work applying AEC to this problem
\cite{Zhao2012,Lonini2013,Lonini2013a}.
did not address how information from the fovea and periphery could be integrated differently at different stages of vergence.

Here, we present a hierarchical mechanism inspired by \cite{Sutton1999,Barto2003,Dietterich2000}
that addresses this problem.
The use of multiple scales has been used in
Different regions of the visual field
first generate different actions.
One of these commands is then selected
based upon the visual information.
The usage of information in the fovea and periphery changes dynamically in a stimulus dependent manner.
Our experimental results demonstrate that this leads to more robust vergence in complex environments.

\section{Models}

\begin{figure}[t]
\includegraphics[width=\textwidth, trim={0 1.5cm 0 0}, clip]{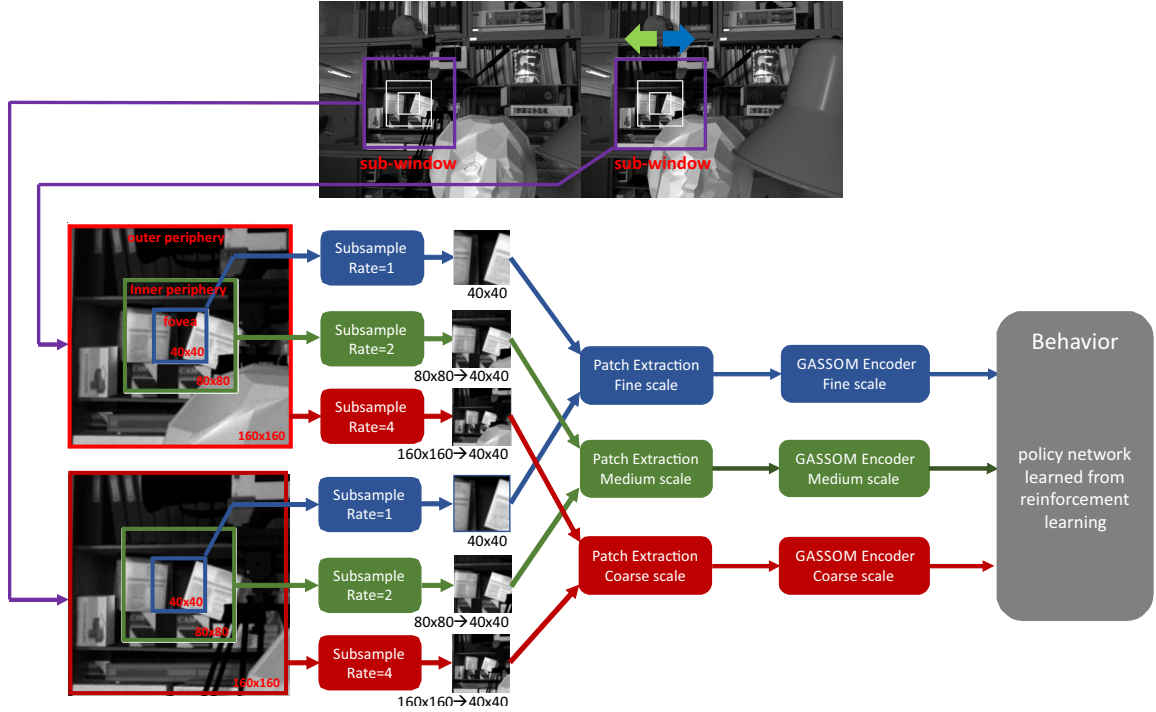}
\caption{Architecture of the vergence joint development system.} \label{fig1}
\end{figure}

Fig. \ref{fig1} shows a block diagram of
our system for the joint development of disparity perception
and fusional vergence behavior.
It consists of three components:
(1) the pre-processing component, which
extracts stereo image patches from sub-windows of the original stereo-images;
(2) the perceptual
component, which encodes the stereo patches as the responses
of a set of disparity selective units using a GASSOM encoder \cite{Chandrapala2015};
(3) the behavior component, which maps the output
of perceptual component to a vergence action via a neural network.
The perceptual and behavioral components are learned simultaneously
as the agent behaves in the environment, through unsupervised and reinforcement learning respectively.

\subsection{Patch extraction}

\begin{figure}[t]
\begin{center}
\includegraphics[width=0.25\textwidth]{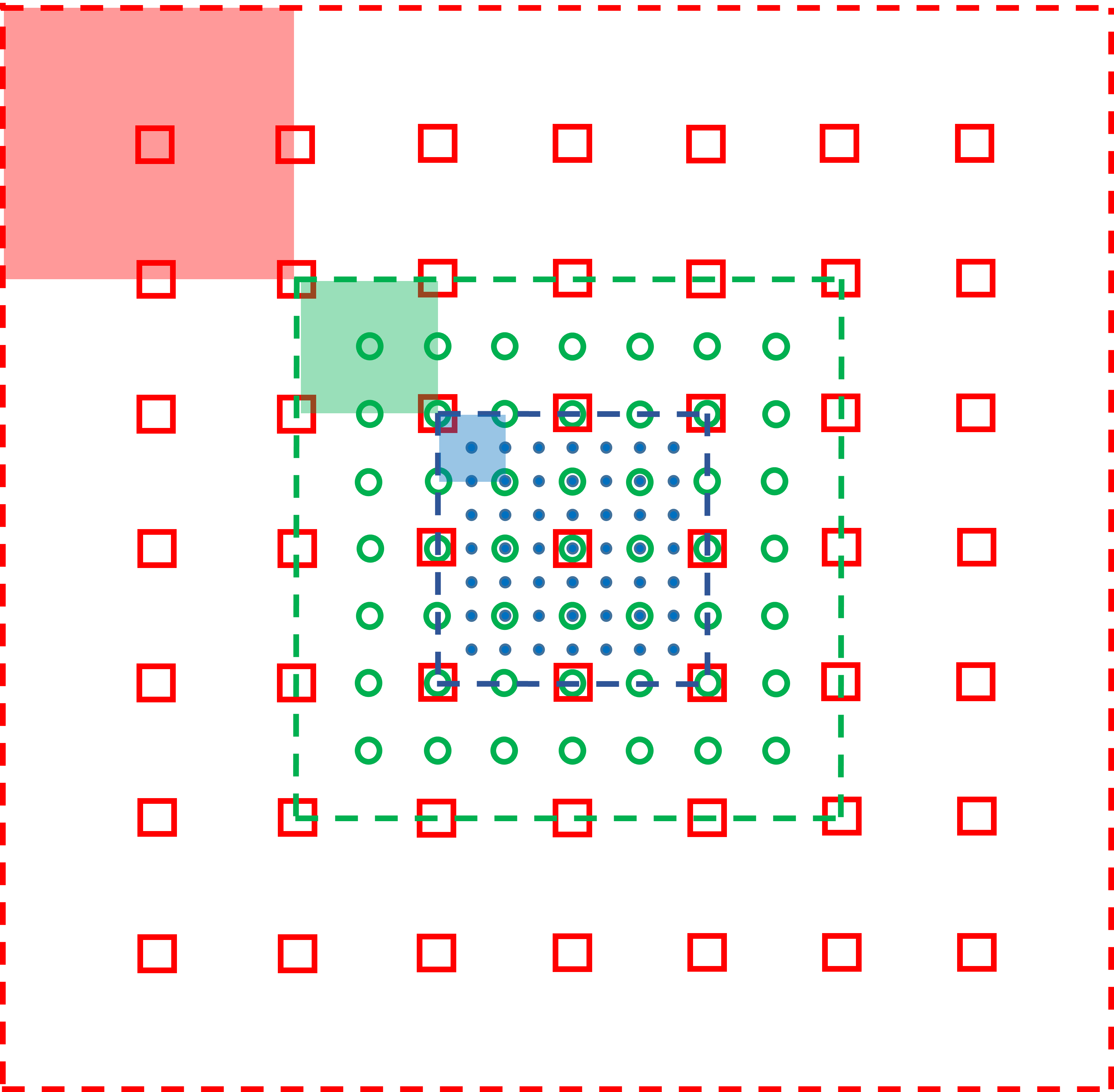}
\end{center}
\caption{
The pyramid structure of the subwindows and patches.
The dotted lines show boundaries of the subwindows
for the fine (blue), medium (green) and coarse (red) scale subwindows.
The solid squares show the size of the region
in the original image covered by a single patch.
The empty squares show the center locations
of the patches.}
\label{fig2}
\end{figure}

We extract three subwindows from each image:
a coarse scale subwindow, a medium scale subwindow and a fine scale subwindow,
Following a pyramidal architecture,
the coarse (medium) scale subwindow is four (two) times the size of the fine scale subwindow, but is downsampled by a factor of four (two), so that all subwindows have the same size (40 by 40 pixels).
These subwindows are further subdivided into a 7 by 7 array of 10 by 10 pixel patches
with a stride of 5 pixels.
As shown in Fig. \ref{fig2},
patches from the different scales
correspond to different sized regions in the original image.

We index scale by $s \in \{\text{f, m, c}\}$
(fine, medium, coarse)
and patch location by $i, j \in \{-3, \ldots, 0, \ldots , 3\}.$
We create stereo image patches $\mathbf{x}_{s,i,j} \in \mathbb{R}^{200 \times 1}$
by combining patches from the two eyes.
\begin{equation}
\mathbf{x}_{s,i,j} =
\begin{bmatrix}
\mathbf{x}_{s,i,j}^\text{L} \\ \mathbf{x}_{s,i,j}^\text{R}
\end{bmatrix}
\end{equation}
\noindent where
$\mathbf{x}_{s,i,j}^\text{L},
\mathbf{x}_{s,i,j}^\text{R} \in
\mathbb{R}^{100 \times 1}$ represent left and right monocular patches.

For intensity and contrast invariance,
we apply mean subtraction
so that $\mathbf{x}_{s,i,j}^\text{L}$ and $\mathbf{x}_{s,i,j}^{\text{R}}$
individually have zero mean,
followed by normalization so that $\mathbf{x}_{s,i,j}$ has unit variance.

\subsection{Perception}

We use the Generative Adaptive Subspace Self Organizing
Map (GASSOM) \cite{Chandrapala2015} to generate a perceptual representation
of the stereo image patches.
The GASSOM algorithm assumes that the high ($N = 200$) dimensional visual input
is generated by sampling from a lower ($B = 2$) dimensional subspace
chosen from a dictionary of $K = 324$ subspaces.
The subspaces are learned through exposure
to unlabeled image patches via an unsupervised learning algorithm.
The learned subspaces represent patches with different oriented textures and stereo disparities.

Patches in the same scale $s$ share the same dictionary.
Each subspace in the dictionary is spanned by a pair of orthogonal
basis vectors contained in the columns of the matrix
$
\mathbf{B}_{s,k} =
\begin{bmatrix}
\mathbf{b}_{s,k,1} & \mathbf{b}_{s,k,2}
\end{bmatrix}
\in \mathbb{R}^{200 \times 2}
$
where $k \in \{0,1,\ldots,323\}$ indexes the dictionary elements.

The squared length of a binocular input patch’s
projection onto the subspace spanned by $\mathbf{B}_{s,k}$
is a measure of the extent to which the subspace can account for
the input patch.
It is also similar to the disparity energy \cite{Ohzawa1990},
which is often used to model the output of disparity selective binocular complex cells.
The disparity is the sum of the squared outputs of two linear binocular neurons.
Because the two columns are orthogonal,
the squared length of the projection can be calculated by
$
|| \mathbf{B}_{s,k}^T \mathbf{x}_{s,i,j} ||
=
(\mathbf{b}_{s,k,1}^T \mathbf{x}_{s,i,j})^2
+
(\mathbf{b}_{s,k,2}^T \mathbf{x}_{s,i,j})^2
$.
Thus, each basis vector is analogous to the receptive field
of a linear binocular neuron.
Each basis vector can be split into two parts,
e.g. $\mathbf{b}_{s,k,1}^\text{L},
\mathbf{b}_{s,k,1}^\text{R} \in \mathbb{R}^{100}$,
one corresponding to the left eye and one corresponding to the right eye.
Each part can be rearranged into a $10 \times 10$ matrix.

The basis vectors are learned
as the agent behaves in the environment by
an unsupervised learning procedure \cite{Chandrapala2015}.
The basis vectors of each subspace $k$
in the direction that minimizes the
reconstruction error of the subspace,
$|| \mathbf{x}_{s,i,j} -
\mathbf{B}_{s,k} \mathbf{B}_{s,k}^\text{T}
\mathbf{x}_{s,i,j} ||^2$.
The reconstruction error is a the squared distance between the input vector and its projection onto the subspace.
The size of the update depends upon how likely the
subspace accounts for the input vector.

After learning,
the basis vectors
develop so that the $10 \times 10$ matrices have Gabor-like profiles
with similar spatial frequencies and orientations \cite{Zhao2012,Lonini2013}.
For each eye, two basis functions,
e.g. $\mathbf{b}_{s,k,1}^\text{L}$ and $\mathbf{b}_{s,k,2}^\text{L}$
are in approximate phase quadrature.
For each basis vector,
the left and right eye components,
e.g. $\mathbf{b}_{s,k,1}^\text{L}$ and $\mathbf{b}_{s,k,1}^\text{R}$
have a phase shift
that determines the preferred disparity of the subspace.
Thus, the learned basis vectors have properties similar to
the linear simple cell binocular receptive fields
in the disparity energy model.

For each scale $s$ and each patch $i,j$,
we define an output feature vector $\mathbf{c}_{s,i,j} \in \mathbb{R}^{324}$
as the set of squared projections onto all subspaces:
\begin{equation} \label{eq:pop}
\mathbf{c}_{s,i,j} =
\begin{bmatrix}
|| \mathbf{B}_{s,0}^T
\mathbf{x}_{s,i,j} ||^2
\\
\vdots
\\
|| \mathbf{B}_{s,324}^T
\mathbf{x}_{s,i,j} ||^2
\end{bmatrix}
\end{equation}
This feature vector models the population output of a set of disparity, spatial frequency and orientation selective complex cells in the primary visual cortex all serving the same location.

\subsection{Behavior}

Behavior is defined by a policy
that maps the output of the perceptual representation
to a vergence action,
We simulate vergence by changing
the horizontal shift between the center locations
of the subwindows extracted from the left and right eye images.
The vergence actions used here update this shift by an amount
$A \in \{-16, -8, -4, -2, -1, 0, 1, 2, 4, 8, 16 \}$
in pixels.

Below, we describe two neural network based
architectures for mapping the perceptual representations
to vergence actions:
a parallel model, which is the same as used in prior work \cite{Zhao2012,Lonini2013}
and a new hierarchical model.

\begin{figure}[t]
\begin{center}
\includegraphics[width=0.8\textwidth]{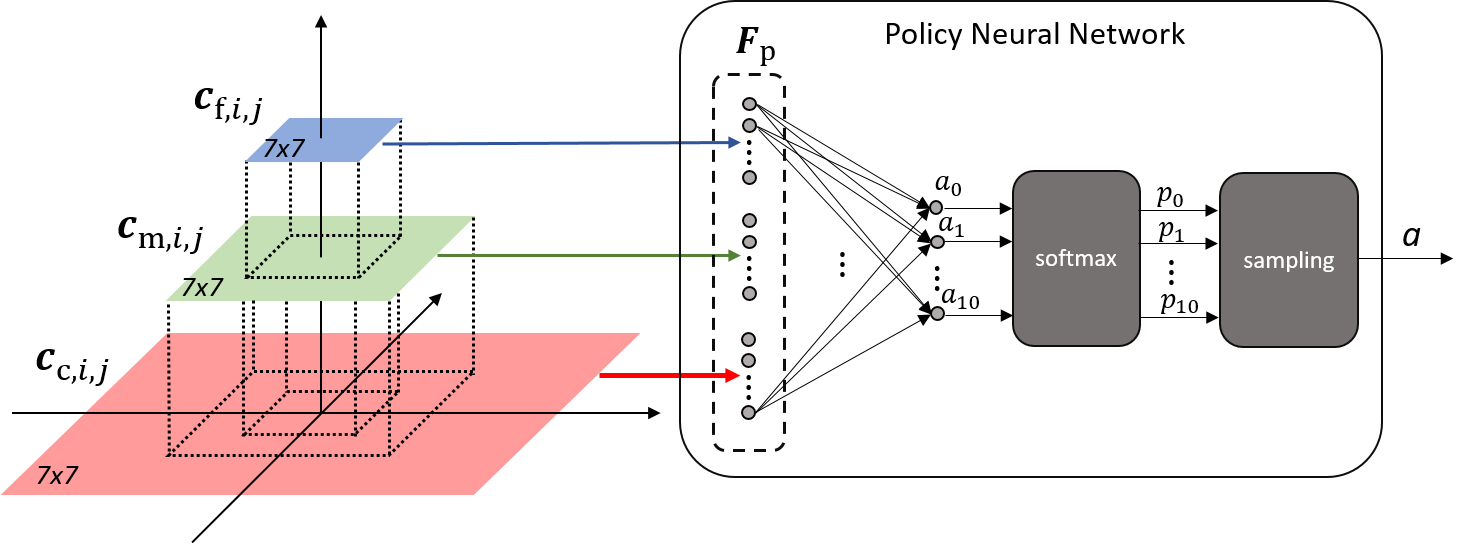}
\\
(a)
\\
\includegraphics[width=0.8\textwidth]{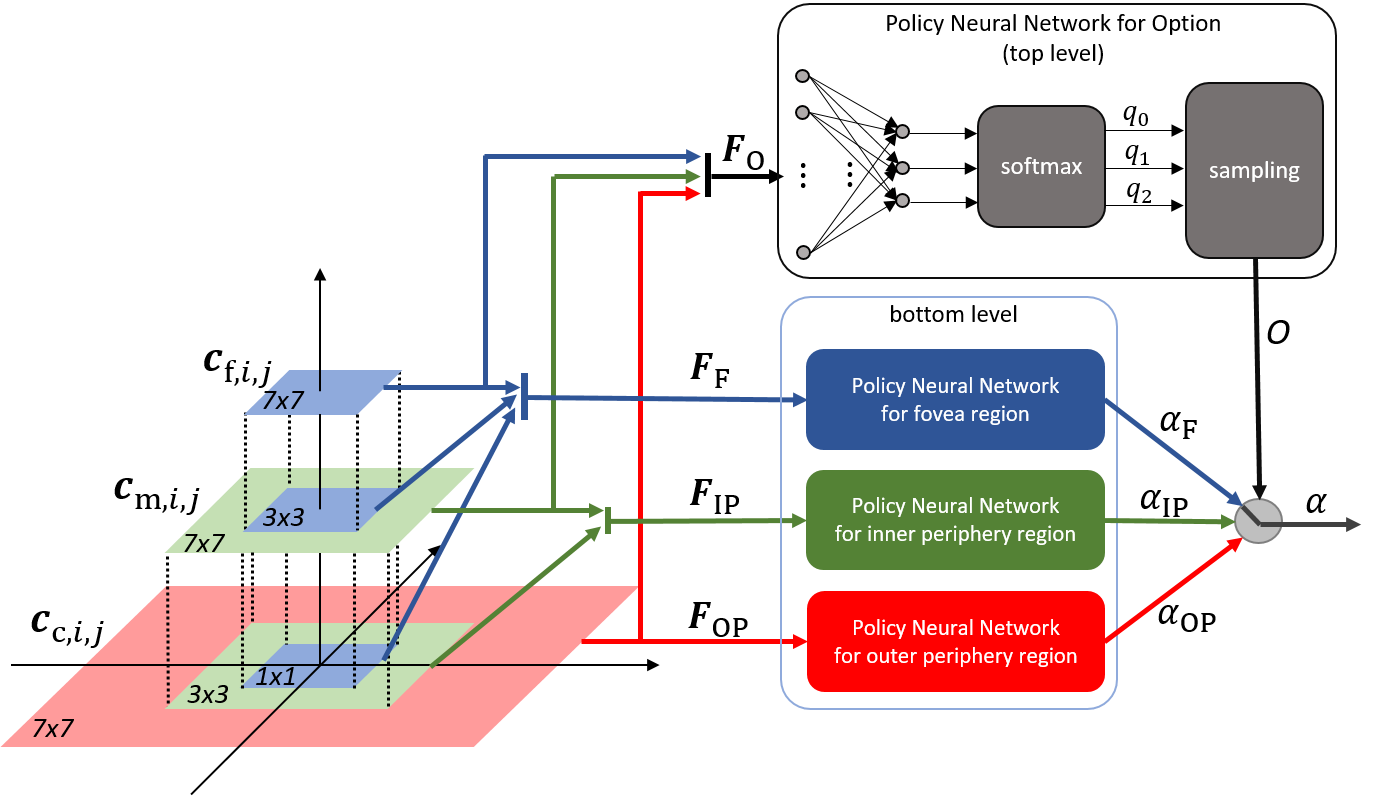}
\\
(b)
\end{center}
\caption{
(a) Architecture of the parallel model.
(b) Architecture of the hierarchical model.}\label{model}
\end{figure}

\subsubsection{Parallel model}

The parallel model (shown in Fig. \ref{model}(a))
combines the population outputs
$\mathbf{c}_{s,i,j} \in \mathbb{R}^{324}$ in (\ref{eq:pop})
into one long feature vector $\mathbf{F}_\text{P}\in \mathbb{R} ^{972}$,
by pooling across space and concatenating across scale
\begin{equation} \label{parallel F}
\mathbf{F}_\text{P}=\begin{bmatrix}
\frac{1}{7^2}\sum_{i,j=-3}^{3}\mathbf{c}_{\text{c},i,j}\\
\frac{1}{7^2}\sum_{i,j=-3}^{3}\mathbf{c}_{\text{m},i,j}\\
\frac{1}{7^2}\sum_{i,j=-3}^{3}\mathbf{c}_{\text{f},i,j}
\end{bmatrix}
\end{equation}
A single layer neural network
with a softmax output nonlinearity
maps this feature vector
to a probability distribution over the 11 actions,
$\mathbf{p} = [ p_i ]_{i=0}^{10} \in \mathbb{R}^{11}$:
\begin{equation} \label{eq:prob}
\mathbf{p} =\text{softmax}(\mathbf{a})
\end{equation}
\noindent where
$\mathbf{a} = [ a_i ]_{i=0}^{10} \in \mathbb{R} ^{11}$
is a set of motor neuron outputs computed by
\begin{equation} \label{eq:motor}
\mathbf{a} = \mathbf{W}_\text{P}\cdot \mathbf{F}_{\text{P}}
\end{equation}
\noindent where $\mathbf{W}_\text{P} \in \mathbb{R} ^{11\times 972}$.
The softmax operator is given by
$p_i={e^{a_i/T}}({\sum_{j=0}^{N}e^{a_j/T}})^{-1}$,
where $T$ is a temperature parameter.

The weights $\mathbf{W}_\text{P}$
develop following the
Natural Actor-Critic Reinforcement Learning (NACREL) algorithm \cite{Bhatnagar2009}
to maximize the discounted sum of instantaneous rewards.
The instantaneous reward is the
reconstruction error of best matching
subspace $e_{s,i,j}$ averaged across all scales and all rewards
\begin{equation}
r_\text{P}=-\frac{1}{3\times 7^2}
\sum_{s}
\sum_{i,j} e_{s,i,j}
\end{equation}
\noindent where
\begin{equation}
e_{s,i,j} =
|| \mathbf{x}_{s,i,j} -
\mathbf{B}_{s,\hat{k}}
\mathbf{B}_{s,\hat{k}}^\text{T}
\mathbf{x}_{s,i,j} ||^2
\end{equation}
\noindent and
$\hat{k} = {\text{argmax}}_{k}
||\mathbf{B}^T_{s,k}
\mathbf{x}_{s,i,j} ||^2$.
The summation ranges over
$s \in \{ \text{f,m,c} \}$
and
$i,j \in \{ -3,\ldots,3 \}$.

Note that both the perceptual component and the behavioral component develop so as to minimize the reconstruction error. This shared goal ensures the stability and robustness of a learning process where both perception and behavior are evolving simultaneously.

\subsubsection{Hierarchical model}

While the parallel model works well with large planar objects,
it does not perform well in more realistic environments,
where the objects being fixated upon do not cover the entire
extent of the fine, medium and coarse scale subwindows.
To address this problem,
we propose here a two level hierarchical model shown in Fig. \ref{model}B.
The bottom level generates three separate vergence commands
based on information from different subwindows.
The top level selects one of these commands.

At the bottom level,
the system defines three separate input feature vectors,
which we label as foveal (F),
inner peripheral (IP), and outer peripheral (OP).
These inputs gather inputs from the image regions
covered by the fine, medium and coarse scale subwindows
respectively,
but may contain informations from multiple scales.
Information within each scale is combined by spatial pooling.
The foveal input,
$\mathbf{F}_\text{F} \in \mathbb{R}^{972}$,
depends not only the responses from all fine scale patches,
but also on the responses from the three by three
array of medium scale patches and the single coarse scale patch
that fall entirely inside the fine scale subwindow.
The inner peripheral input,
$\mathbf{F}_\text{IP}\in \mathbb{R}^{648}$,
depends on both medium and coarse scale patches.
The outer peripheral input,
$\mathbf{F}_\text{OP}\in \mathbb{R}^{324}$,
depends only coarse scale input.
\begin{equation}  \label{hiera foveal input}
\mathbf{F}_\text{F}=\begin{bmatrix}
\mathbf{c}_{\text{c},0,0}\\
\frac{1}{3^2}\sum_{i,j=-1}^{1}\mathbf{c}_{\text{m},i,j}\\
\frac{1}{7^2}\sum_{i,j=-3}^{3}\mathbf{c}_{\text{f},i,j}
\end{bmatrix}
\hspace{0.5cm}
\mathbf{F}_\text{IP}=\begin{bmatrix}
\frac{1}{3^2}\sum_{i,j=-1}^{1}\mathbf{c}_{\text{c},i,j}\\
\frac{1}{7^2}\sum_{i,j=-3}^{3}\mathbf{c}_{\text{m},i,j}
\end{bmatrix}
\hspace{0.5cm}
\mathbf{F}_\text{OP}=\frac{1}{7^2}\sum_{i,j=-3}^{3}\mathbf{c}_{\text{c},i,j}
\end{equation}

Each feature vector generates
a different probability distribution over actions
following a similar strategy as used in the parallel model,
i.e. equations (\ref{eq:prob}) and (\ref{eq:motor})
with appropriate changes in the dimensionality of
the input feature vector and weight matrix.

Weights were learned using the NACREL algorithm.
The instantaneous reward for each networks was
the average reconstruction
errors of the patches included in its feature vector.
\begin{equation} \label{hiera fovea reward}
r_\text{F}=-\frac{1}{3}
\left(
e_{\text{c},0,0}+
\frac{1}{3^2}\sum\nolimits_{i,j=-1}^{1}e_{\text{m},i,j}+
\frac{1}{7^2}\sum\nolimits_{i,j=-3}^{3}e_{\text{f},i,j}
\right)
\end{equation}
\begin{equation} \label{hiera innerP reward}
r_\text{IP}=-\frac{1}{2}
\left(
\frac{1}{3^2}\sum\nolimits_{i,j=-1}^{1}e_{\text{c},i,j}+
\frac{1}{7^2}\sum\nolimits_{i,j=-3}^{3}e_{\text{m},i,j}
\right)
\hspace{0.2cm}
r_\text{OP}=-\frac{1}{7^2}
\sum\nolimits_{i,j=-3}^{3}e_{\text{c},i,j}
\end{equation}

The top level,
selects from
among the three options based on
the pooled information
from all scales,
$\mathbf{F}_{\text{O}}=\mathbf{F}_\text{P}$,
using a neural network with the same structure
as the parallel network,
except the dimensionality of the output vector
was three.
Each output corresponds to the selection of
the vergence command from one of the
bottom layer networks.
The weights optimized the
same reward as the parallel model,
$r_{\text{O}}=r_\text{P}$.

%

\section{Experimental Procedure}
\label{ExpProc}

A simulated agent was presented with a sequence of virtual scenes.
Within each scene, the agent executed saccades
to 20 different fixation locations,
chosen randomly by sampling
from the saliency distribution
computed on the left eye image
using Attention based on Information Maximization (AIM) \cite{Bruce2009}.
At each fixation location,
the agent executed 10 vergence commands.
After 20 fixations, a new scene was presented to the agent.

During training,
vergence actions $\alpha$
were generated by sampling from the
multinomial probability distribution specified by $\mathbf{p}$.
During testing,
actions were generated using a greedy policy,
which chose the most likely action.

We mimic the effect of eye movements in a scene
by changing the center locations of the subwindows
taken from a single stereo image pair
from the Tsukuba Stereo Dataset \cite{Martull2012}.
This dataset includes 1800 stereo pairs
generated by rendering the video
obtained by a parallel stereo camera pair
moving through a simulated laboratory environment
and corresponding ground truth disparity maps.
Since adjacent frames in the video are very similar,
we took one out of every five frames in the sequence,
resulting in 360 stereo images.
Twenty were selected randomly as the testing set.
The remainder were used as the training set.

Fixation/vergence actions were executed by
setting the the center location
of the left eye subwindows
equal to the chosen fixation point.
The center position of the right eye subwindows
was offset horizontally by the vergence,
which was updated according the vergence actions chosen.

Due to the parallel camera geometry of the Tsukuba dataset
and the way we simulated vergence movements,
the system only encounters horizontal disparities.
In reality, a convergent camera geometry will introduce vertical disparities due to shifts in the epipolar lines.
However, we do not expect this additional complexity
to dramatically alter the results presented here.
The AEC framework
leads to similar results for vergence control
using eye movements simulated by window movement \cite{Zhao2012}
and for camera movements in simulated and real environments \cite{Lonini2013a}.

\section{Experimental Results}

\subsection{Trained Policies}

\begin{figure}[!t]
\begin{center}
\includegraphics[width=0.9\textwidth]{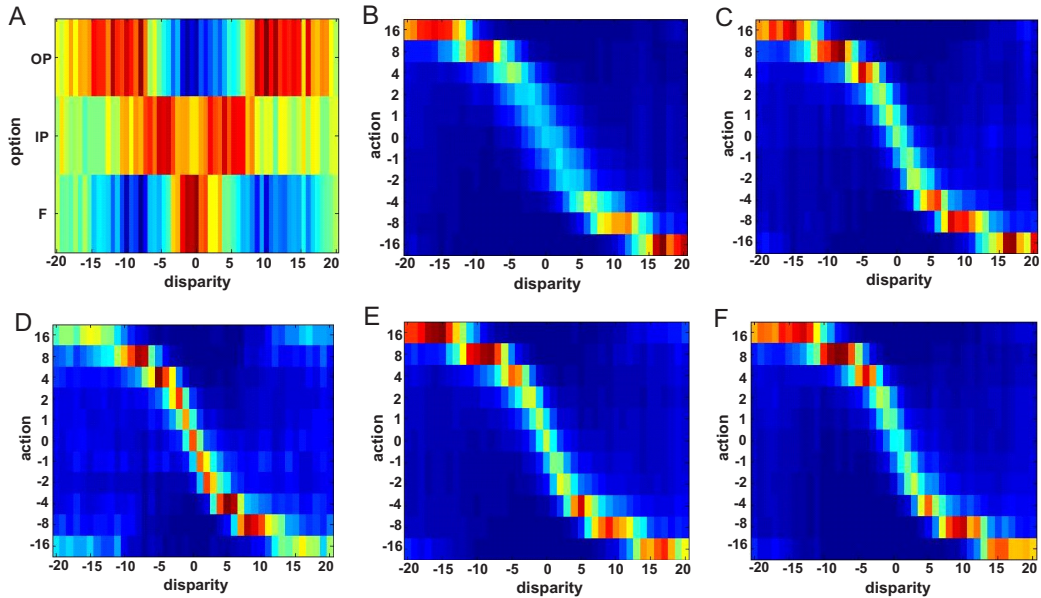}
\caption{
Learned policies from one trial.
Each column of an image
shows the probability of an action
given the input disparity
using the jet colormap,
which varies from blue (0) to red (1).
(A-E) Hierarchical model: (A) top level,
(B) outer peripheral,
(C) inner peripheral,
(D) foveal,
(E) combined.
(F) Parallel model.} \label{policies}
\end{center}
\end{figure}

Fig. \ref{policies} visualizes the policies learned during one trial
of the hierarchical and parallel models as images.
Each column
shows the value of $\mathbf{p}$ in (\ref{eq:prob})
at a particular disparity $d$
averaged over 100 binocular inputs
generated from randomly selected left eye images
in the test set.
Left eye subwindows were taken around a
randomly generated center location.
Right eye subwindows were taken from the same image
around a center location offset by $d$.
This ensured that the disparity is
uniform across the subwindows.

The learned policies have properties that reflect
and appropriately exploit
the properties of the different inputs.

The top level policy of the hierarchical model (Fig. \ref{policies}A)
tends to choose the action generated by the outer peripheral policy when the disparity is large,
by the foveal policy when the disparity is small,
and the inner peripheral policy for intermediate disparities,
leading to a "V" shaped image.

For the vergence action generating policies (Fig. \ref{policies}B-F),
we expect to see an upside-down sigmoid
due to the exponential action spacing.
Positive input disparities generate
negative vergence actions to zero-out the disparity.
For the hierarchical model,
the outer peripheral policy (B)
reliably generates large vergence actions for large input disparities,
but is less reliable at small input disparities,
due to the coarser resolution.
In contrast, the foveal policy (D)
more reliably generates actions that precisely cancel the input disparity when it is small, but is less reliable at larger disparities, due to the limited spatial extent of the patches (10 by 10 pixels).
When the input disparity is
near -20 or +20 pixels,
we observe a bimodal distribution with centers
at the -10 and +10 vergence actions.
The left and right eye patches
image non-overlapping regions in the environment.
The policy reliably detects
that the input disparity is large,
but is unsure about the direction to change the vergence.
The inner peripheral policy (C) exhibits intermediate characteristics.

We estimated the policy combining the top and bottom levels
by averaging each column of the three vergence action policies
weighted by the probability that each policy was selected at the top level.
The combined policy (E) is very similar to the policy from the parallel method (F), as we might expect for these inputs where the
disparity is uniform over all subwindows.

\subsection{Vergence in Complex Environments}

\begin{figure}[!t]
\begin{center}
\includegraphics[width=3.3in, trim={0 0 0 0}, clip] {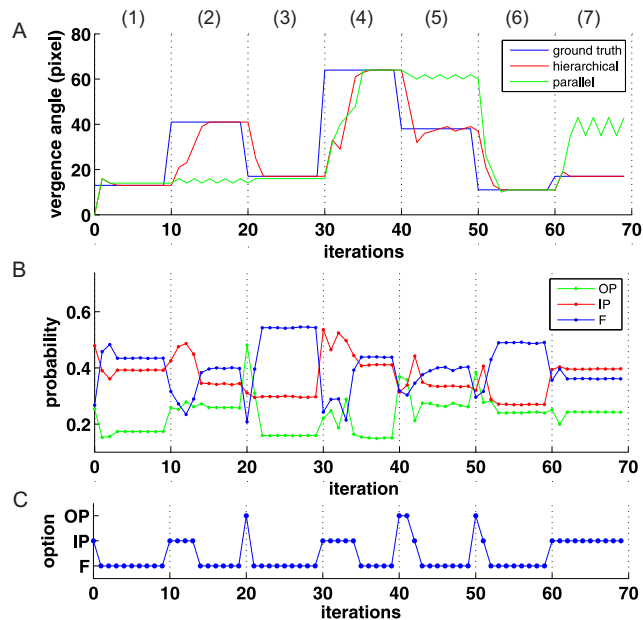}
\caption{
Model outputs over 7 fixations.
(A) The vergence angle trajectories generated by the hierarchical (red) and parallel (green) models in comparison to
the ground truth (blue) vergence angles required to achieve zero retinal disparity at the subwindow center.
(B) The probabilities for selecting the actions
generated from the
outer (green) and inner (red) peripheral and foveal (blue) networks,
as computed by the top level network.
(C) The option chosen by the top level network.
Dotted gray lines delinate fixations,
which are numbered for reference at the top.
} \label{compare_example_traces}
\end{center}
\end{figure}

The difference between the parallel and hierarchical model
is clearly evident in more complex environments
where the disparities encountered in the fovea and periphery
may differ, e.g. when the agent is fixating on a small object
that does not cover the entire periphery and the background
is at a different depth.

Fig. \ref{compare_example_traces}A shows the vergence
angle trajectories generated by the parallel and hierarchical models
for a sequence of seven fixations
on one of the stereo images in the test set.
The parallel model failed to converge to the correct vergence angle
at the $2^{nd}$, $5^{th}$ and $7^{th}$ fixations,
and exhibited oscillatory behavior due to conflicting disparities
in the fovea and periphery..
The hierarchical model performed much better,
converging without oscillation on all fixations.

Fig. \ref{compare_example_traces}B
shows that at the beginning of the fixations,
when the retinal disparities are large,
the hierarchical model tends to choose
actions generated by the peripheral regions.
At the end of the fixations,
when the retinal disparity at the fixation point is small,
it chooses actions generated by the fovea.
A clear progression from outer periphery to fovea
can be seen in the $5^{th}$ and $6^{th}$ fixations.

\section{Conclusion}

We have proposed a computational architecture
for the joint learning of disparity perception and vergence control,
which incorporates a hierarchical model for integrating
information from the fovea and periphery.
This architecture enables an agent to learn how to to resolve
conflicting disparities in different image regions
through its interaction with the environment,
and without an explicit teaching signal.
The policy that emerges exhibits behavior
reminiscent of coarse to fine behavior.

\vspace{0.5cm}

\noindent \textbf{Acknowledgements.} This work was supported by
the Hong Kong Research
Grants Council under Grant 16244416,
the German Federal Ministry of
Education and Research under Grants 01GQ1414 and 01EW1603A, the European Union's Horizon 2020 Grant  713010, and the Quandt Foundation.

\bibliography{mybibliography}

\end{document}